\documentclass[11pt]{article}

\PassOptionsToPackage{numbers}{natbib}
\usepackage[preprint]{acl}

\usepackage{times}
\usepackage{latexsym}

\usepackage[T1]{fontenc}

\usepackage[utf8]{inputenc}

\usepackage{microtype}

\usepackage{inconsolata}

\usepackage{graphicx}

\usepackage{hyperref}       
\usepackage{url}            
\usepackage{booktabs}       
\usepackage{amsfonts}       
\usepackage{amsmath}       
\usepackage{nicefrac}       
\usepackage{microtype}      
\usepackage[dvipsnames]{xcolor}
\usepackage[inline]{enumitem}
\usepackage{makecell}
\usepackage{fancyhdr}       
\usepackage{subfig}
\usepackage{mdframed}
\usepackage{comment}
\usepackage{tabularx}
\usepackage{caption}
\usepackage{placeins}
\usepackage{float}
\usepackage{array} 
\usepackage{wrapfig}
\usepackage{cleveref}
\usepackage{multirow}
\usepackage{algorithm}
\usepackage{algorithmic}
\usepackage{listings}
\usepackage{atbeginend}
\usepackage{csquotes} 
\MakeOuterQuote{"}
\usepackage{lipsum}
\usepackage{appendix}

\AtBeginEnvironment{appendices}{\crefalias{section}{appendix}}

\newenvironment{packed_enum}{
\begin{enumerate}
	\setlength{\itemsep}{1pt}
	\setlength{\parskip}{0pt}
	\setlength{\parsep}{0pt}
}{\end{enumerate}}

\AfterBegin{packed_enum}{\vspace{-0.7em}}
\AfterEnd{packed_enum}{\vspace{-0.7em}}
\AfterBegin{packed_item}{\vspace{-0.7em}}
\AfterEnd{packed_item}{\vspace{-0.7em}}
\AfterBegin{packed_description}{\vspace{-0.7em}}
\AfterEnd{packed_description}{\vspace{-0.7em}}

\AfterBegin{quote}{\vspace{-0.7em}}
\AfterEnd{quote}{\vspace{-0.7em}}

\lstdefinelanguage{json}{
basicstyle=\ttfamily,
numbers=left,
numberstyle=\tiny\color{gray},
stepnumber=1,
numbersep=8pt,
showstringspaces=false,
breaklines=true,
literate=
*{0}{{{\color{blue}0}}}{1}
{1}{{{\color{blue}1}}}{1}
{2}{{{\color{blue}2}}}{1}
{3}{{{\color{blue}3}}}{1}
{4}{{{\color{blue}4}}}{1}
{5}{{{\color{blue}5}}}{1}
{6}{{{\color{blue}6}}}{1}
{7}{{{\color{blue}7}}}{1}
{8}{{{\color{blue}8}}}{1}
{9}{{{\color{blue}9}}}{1}
{:}{{{\color{red}:}}}{1}
{,}{{{\color{red},}}}{1}
{\{}{{{\color{orange}\{}}}{1}
{\}}{{{\color{orange}\}}}}{1}
{[}{{{\color{orange}[}}}{1}
{]}{{{\color{orange}]}}}{1},
}

\lstdefinelanguage{txt}{
basicstyle=\ttfamily,
numbers=left,
numberstyle=\tiny\color{gray},
stepnumber=1,
numbersep=8pt,
showstringspaces=false,
breaklines=true,
}

\usepackage{newfloat}
\floatstyle{ruled}
\newfloat{listing}{tb}{lst}{}
\floatname{listing}{Listing}
\title{Query Disambiguation via Answer-Free Context:\\Doubling Performance on Humanity’s Last Exam}

\author{%
Michael Majurski$^{1,2*}$ \quad Cynthia Matuszek$^{2}$\\
$^1$National Institute of Standards and Technology \quad $^2$University of Maryland Baltimore County\\
\texttt{michael.majurski@nist.gov}\\
\texttt{cmat@umbc.edu}\\
}

\begin{document}
\maketitle

\begin{abstract}
	How carefully and unambiguously a question is phrased has a profound impact on the quality of the response, for Language Models (LMs) as well as people.
	While model capabilities continue to advance, the interplay between grounding context and query formulation remains under-explored.
	This work investigates how the quality of background grounding information in a model's context window affects accuracy. 
	We find that combining well-grounded dynamic context construction (i.e, RAG) with query rewriting reduces question ambiguity, resulting in significant accuracy gains.
	Given a user question with associated answer-free grounding context, rewriting the question to reduce ambiguity produces benchmark improvements without changing the answer itself, even compared to prepending that context before the question. Using \texttt{gpt-oss-20b} to rewrite a subset of Humanity's Last Exam using answer-free grounding context improves \texttt{gpt-5-mini} accuracy from 0.14 to 0.37.
	We demonstrate that this accuracy improvement cannot be fully recovered just through prompting at inference time; rather, distinct rewriting and answering phases are required.
	Code and data are available at \url{https://github.com/mmajurski/lm-rewrite-uplift}
\end{abstract}

\section{Introduction}


The ongoing explosion of Language Model (LM) capability is largely a consequence of increases in training compute, model parameters, and dataset size as described by scaling laws~\citep{chen2025scaling}; which have demonstrated relationships between performance and parameter count, dataset size, and computation. 
This capability growth has opened a widening chasm between what modern LMs can demonstrably do and how we measure their spiky intelligence. 
%
%
Leaderboard-style benchmarks have been a primary engine and evaluator of progress in machine learning, providing an objective scalable way to measure and compare capabilities. 
However, human-curated static benchmarking is a brittle standard.
\begin{figure}[t!]
	\centering
	\includegraphics[width=0.9\columnwidth]{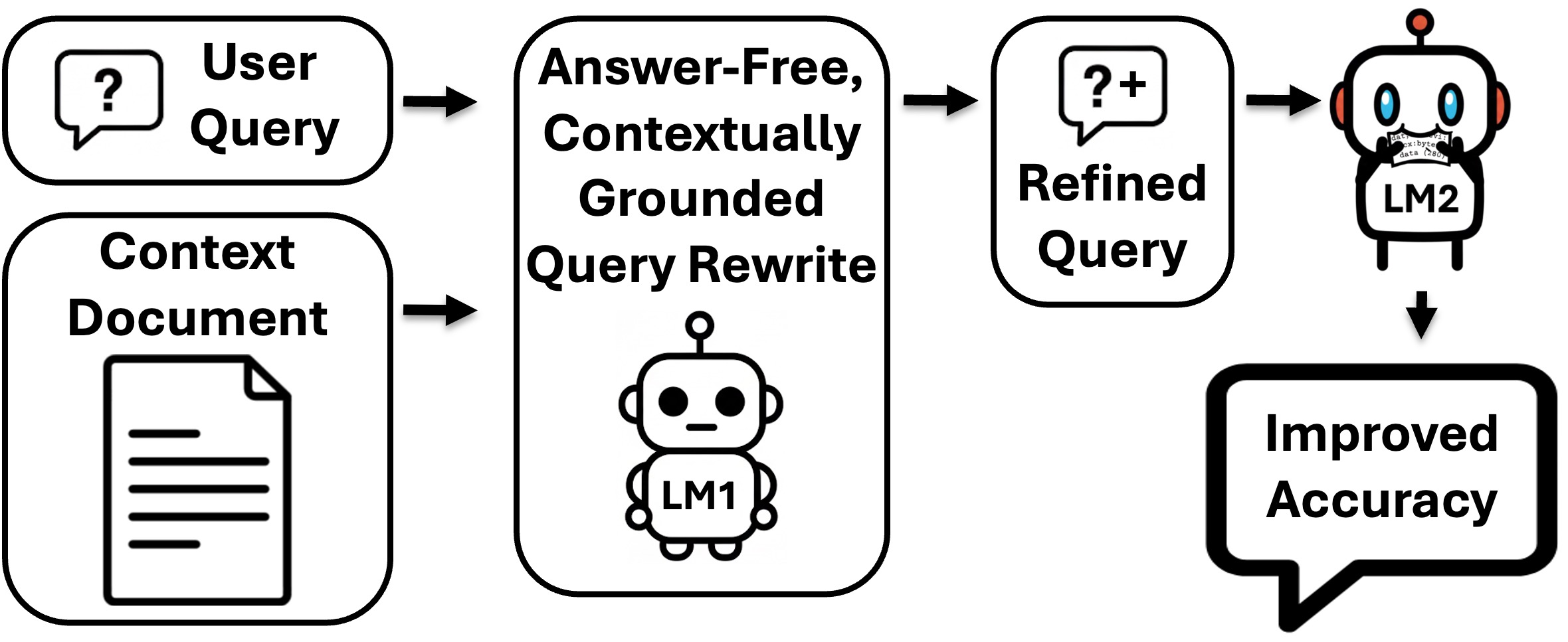}
	\vspace{-0.25em}
	\caption{When RAG systems surface relevant information, LM performance can be enhanced by rewriting the initial query using \textit{context}---added information that, without providing the answer, gives relevant background knowledge and direction.}
	\label{fig:architecture}
	\vspace{-1em}   
\end{figure}
\begin{figure}[b!]
	\vspace*{-1.0em}
	\centering
	\includegraphics[width=0.95\columnwidth]{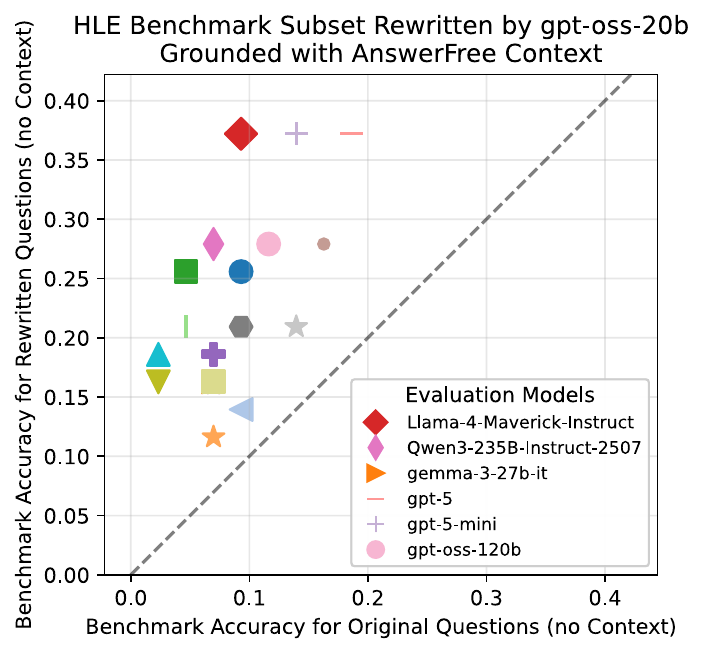}
	\vspace{-0.5em}
	\caption{Rewriting questions 
		using answer-free grounding context yields significant accuracy improvement (distance above line) over the original questions, as evaluated on a subset of Humanity's Last Exam (HLE). See \Cref{fig:all-scatterplots} for a complete legend.
	}
	\label{fig:hle-uplift}
\end{figure}
LM users often implicitly assume the LM shares their mental model, including their background knowledge, context, and intent.
This can cause them to omit critical background and related information when formulating a query, believing it to be self-evident.
The model, lacking this context, operates instead on training data patterns and the explicit text of the prompt. 
When faced with an under-specified query, its response will reflect implicit (or explicit) assumptions.
If these assumptions do not align with the user's unstated background expectations, the resulting answer (while potentially factually correct under that assumed interpretation) will be perceived as incorrect or irrelevant.
We focus on the specific version of this problem where implicit information can be clarified within RAG search systems. 


In this work, we explore a the impact on system performance of adding Answer-Free Context (\texttt{AFC})---background information that is relevant to the question but which does not contain the actual answer.
We demonstrate that even when RAG fails to surface a direct answer, the retrieved background information can be used to perform a contextually grounded disambiguation rewrite (\Cref{fig:architecture}) to improve an underspecified query before it is answered.
This method can meaningfully improve accuracy on the human validated subset of Humanity's Last Exam (HLE) (\Cref{fig:hle-uplift}), producing better results than simply including the \texttt{AFC} in the question.
An example of an original question (from Squadv2~\citep{DBLP:journals/corr/abs-1806-03822}), associated \texttt{AFC}, and rewritten question are shown below. 

\begin{mdframed}
	\footnotesize
	\textbf{Original question:} What kind of lasers are crystals of zinc suflde \textit{(sic)} used in?
	
	\noindent\textbf{Answer-free context:} Zinc chloride is often added to lumber as a fire retardant and can be used as a wood preservative. It is also used to make other chemicals. Zinc methyl (CH\textsubscript{3}Zn) is used in a number of organic syntheses. Zinc sulfide (ZnS) is used in luminescent pigments such as on the hands of clocks, X-ray and television screens, and luminous paints. Crystals of ZnS are used in lasers. Zinc sulfate is a chemical in dyes and pigments. Zinc pyrithione is used in antifouling paints.
	
	\noindent\textbf{Rewritten question:} Which portion of the electromagnetic spectrum do lasers that incorporate zinc sulfide (ZnS) crystals generally operate in?
	
	\noindent\textit{(Prompts are given in the Appendix.)}
\end{mdframed}


\noindent This work makes the following contributions:\vspace{0.5ex}
\begin{packed_enum}
	\item Introduces a method for leveraging Answer-Free Context (\texttt{AFC}) to disambiguate queries, yielding significant accuracy gains.
	\item Analyzes the performance differential in RAG systems when retrieving direct answers versus purely auxiliary background information. 
	\item Demonstrates that this accuracy improvement necessitates a distinct rewriting phase, and cannot be replicated by simply prepending the retrieved context to a prompt. 
\end{packed_enum}

\section{Related Works}


While LMs store significant parametric knowledge, their responses are not grounded in reliable external information sources.
Considerable background information lets these models score highly on tests of general knowledge, but they often struggle with domain-specific or proprietary data absent from public training sets. 
Dynamic context construction approaches like Retrieval-Augmented Generation (RAG) address this by retrieving external evidence during the generation process, which significantly improves factual accuracy~\citep{sharma2025retrieval}.
The RAG pipeline introduces its own optimization challenges, spanning document chunking, search, ranking~\citep{NEURIPS2024_db93ccb6}, and post-retrieval processing~\citep{dai2025evinote}.

A particularly critical component, with a long history in information retrieval, is the transformation, expansion, and normalization of user queries~\citep{Rastogi2019,rivas2014study}.
In professional and enterprise workflows, system performance can depend on RAG to dynamically construct context for the LM based upon user queries~\citep{lewis2020retrieval,NEURIPS2024_db93ccb6}.
However, current RAG evaluation focuses on retrieval ranking~\citep{yang2024crag,NEURIPS2024_27245589} rather than how models \textit{utilize} retrieved documents---information that might be relevant without explicitly containing the answer.

The initial user query is often an imperfect expression of the information needed, suffering from ambiguity, missing context, or poorly aligned terminology with a target document corpus. 
To bridge this lexical gap, modern RAG systems employ query expansion and rewriting~\citep{jagerman2023query,zhou2023unified,li2024dmqr}.
Query expansion reformulates the question to better surface relevant documents and improve the signal-to-noise ratio of the retrieved context~\citep{gao2023retrieval}.
%
LMs have proven adept at this rewriting, expansion, and contextualization task, disambiguating queries using carefully designed prompts~\citep{wilson2025contextualizing,gao2023retrieval,sun2025picos}. 
Other approaches explore post-RAG retrieval optimizations for improving signal to noise in the surfaced documents, shifting from a retrieval-generate paradigm to retrieval-note-generate in which the discovered documents are summarized into high level notes~\citep{dai2025evinote}.

Other dynamic context and prompting strategies focus on getting the most out of the context.
For example, Step-Back Prompting attempts to have LMs derive high-level concepts from first principles to guide reasoning in query answering~\citep{zheng2023take}.
Frameworks like AGREE have the LM cite its sources within the context window (drawn from grounding documents)~\citep{ye2023effective}.
Other approaches involve extensions to chain-of-thought, including Meta-CoT, which models how to determine what underlying reasoning is required to arrive at a particular CoT~\citep{xiang2025towards}.
CoT-type reasoning approaches enable the LM to employ self-directed question rewriting during the process of figuring out how best to respond.
Our work demonstrates that prompting LMs to disambiguate questions in COT is less effective than a separate rewriting phase.

\section{Methodology}

\begin{figure}[t!]
	\vspace*{-0.25em}
	\centering
	\includegraphics[width=0.9\columnwidth]{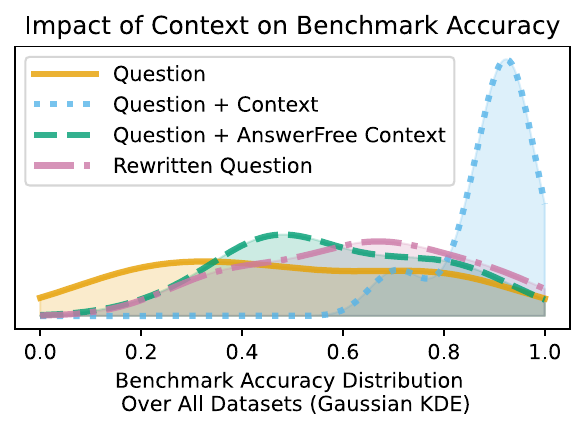}
	\vspace{-0.8em}
	{\small
		\caption{The quality of context presented to an LM has an impact on question-answering performance. Accuracy is highest when RAG systems surface context containing the answer (\textcolor{cyan}{cyan}), but when the question is presented without context (\textcolor{orange}{orange}) or the surfaced information does not contain the answer (\textcolor{ForestGreen}{green}), performance (without query rewriting) suffers. The act of interpretation during question rewriting (\textcolor{Rhodamine}{pink}) produces an accuracy improvement beyond just prepending the \texttt{Question + AnswerFree} \texttt{Context} information used to rewrite before the question; despite not including the \texttt{AFC} when asking an LM the rewritten question.}
		\label{fig:impact-of-context}
	}	
	\vspace{-1.0em} 
\end{figure}

This work explores applying RAG query rewrite/expansion techniques directly to queries posed to LMs, which results in an accuracy improvement on evaluation benchmarks. 
Benchmarks routinely evaluate knowledge by posing factual questions~\citep{wang2024mmlu}. \Cref{fig:impact-of-context} shows the impact of added context on \texttt{gpt-oss-120b}'s performance (across all datasets used in this study)(see \Cref{tab:dataset_publication_cutoff}), and demonstrates that naively adding \texttt{AFC} without rewriting the query to be clearer and less ambiguous does not improve the results as significantly as desired.
However, as shown in \Cref{fig:hle-uplift}, using \texttt{AFC} to rewrite the question can both disambiguate what is being asked and fill in relevant background assumptions, producing robust accuracy gains. 
%
Just prepending the \texttt{AFC} text before the question does not produce an equivalent accuracy improvement, as shown in \Cref{fig:impact-of-context}, where \texttt{Question + AnswerFree} \texttt{Context} under-performs \texttt{Rewritten} \texttt{Question}.
Unless otherwise stated the \texttt{Rewritten} condition never includes the \texttt{AFC}.

We hypothesize that the accuracy improvement is caused by the act of interpretation during rewriting, via clarifying and disambiguating the question separately from attempting to answer it (\Cref{sec:task_seperation} provides evidence for this).
It is worth noting that question rewriting using \texttt{AFC} does not recover the full performance improvement of adding context which directly contains the correct answer.
Prior work on RAG systems has demonstrated the utility of query expansion~\citep{wilson2025contextualizing} and prior work on generative benchmarking demonstrated a correlation between generated question length and model benchmark accuracy~\citep{majurski2025generative}.
Applying rewriting to knowledge benchmarks produces an improvement in accuracy. 
A potential mechanism of action is that the rewritten prompt may shift the embedding space trajectory during generation to more closely align with a latent representation that contains the correct answer.
We explore that trend and characterize how well the effect generalizes.

\subsection{Datasets \& Models}

To explore the impact of question rewriting on LM benchmark performance, datasets are needed that contain pairs of questions with associated context that provides the correct answer. While most RAG systems are designed to produce that type of information, validated reference data is in comparatively short supply, as benchmark creation across a variety of domains is historically an expensive and frequently manually intensive process. Additionally, in order to evaluate whether observed results stem from benchmark question memorization via training data contamination, datasets are needed that contain information that postdates model training.
%
\Cref{tab:dataset_publication_cutoff} lists the datasets used in this study and gives the release data of the benchmark. 
Only Humanity's Last Exam (HLE)~\citep{HumanityLastExam} and generative benchmarking data was post knowledge cutoff for all models tested.

\begin{table}[h!]
	\centering
	\caption{Evaluation Datasets and Publication Dates}
	\scriptsize
	\label{tab:dataset_publication_cutoff}
	\vspace{-1em}   
	
	\begin{tabular}{p{5.5cm} p{1.25cm}}
		\toprule
		\textbf{Dataset} & \textbf{Release Date}  \\
		Public Datasets & \\
		\toprule
		Humanity's Last Exam~\citep{HumanityLastExam} & 2025\\
		\hline
		Squadv2~\citep{DBLP:journals/corr/abs-1806-03822} & 2018 \\
		\hline
		HotpotQA~\citep{yang2018hotpotqa} & 2018 \\
		\hline
		TrivaQA-web~\citep{2017arXivtriviaqa} & 2017 \\
		\hline
		NaturalQuestionsShort~\citep{kwiatkowski2019natural} & 2019 \\
		\hline
		PubMedQA~\citep{jin2019pubmedqa} & 2019 \\
		\hline
		BoolQ~\citep{clark2019boolq} & 2019 \\
		\hline
		FermiQA~\citep{kalyan2021much} & 2021 \\
		\hline
		MS-MARCO-QA~\citep{bajaj2016ms} & 2016 \\
		\hline
		MusiqueQA~\citep{trivedi2022musique} & 2022 \\
		\hline
		2WikiMultiHopQA~\citep{ho2020constructing} & 2020 \\
		\bottomrule
		\vspace{0.1em}   
		Generative Benchmarks (constructed using~\citep{majurski2025generative}) & \\
		\toprule
		arXiv\_2502\_17521v1~\citep{chen2025recent} & 2025 \\
		\hline
		America's AI Action Plan~\citep{AiPlan} & 2025 \\
		\bottomrule
		\vspace{0.1em}   
		Generative Benchmarks (constructed using~\citep{shashidhar2025yourbench}) & \\
		\toprule
		arXiv\_2502\_17521v1~\citep{chen2025recent} & 2025 \\
		\hline
		America's AI Action Plan~\citep{AiPlan} & 2025 \\
		\bottomrule
	\end{tabular}
	\vspace{-1em} 
\end{table}

All datasets (except HLE) contain human curated associated context with each question.
The originally published HLE dataset only contains answer rationale, which is insufficient context.
However, FutureHouse released manual validations for a subset of chemistry/biology questions with grounding literature~\citep{HleFutureHouse}.
This work uses that FutureHouse validated subset of HLE which contains (per question) human vetted evidence and context drawn from published sources.

Evaluating the efficacy of query rewriting for LM benchmark evaluation requires two categories of models.
First is the model performing question rewriting (which may not be the same as the model under evaluation via the benchmark).
We separate the question rewriting and the evaluation phases to ensure that all evaluation models are given the same rewritten questions to reduce measurement noise. 
Second are the set of models evaluated using the rewritten questions; these span a wide range of size and capability.
\Cref{tab:model_cutoff} outlines the models we evaluated on the rewritten question benchmarks.

\begin{table}[h!]
	\centering
	\caption{Evaluation Models Knowledge Cutoff}
	\label{tab:model_cutoff}
	\scriptsize
	\vspace{-1em} 
	
	\begin{tabular}{p{3.6cm} p{1.2cm} p{1.2cm}}
		\toprule
		\textbf{Model} & \textbf{Knowledge-Cutoff} & \textbf{Public-Release}  \\
		\hline
		\texttt{gpt-5} & Sep 2024 & Aug 2025  \\
		\hline
		\texttt{gpt-5-mini} & May 2024 & Aug 2025  \\
		\hline
		\texttt{gpt-5-nano} & May 2024 & Aug 2025  \\
		\hline
		\textbf{\texttt{gpt-oss-20b}} & Jun 2024 & Aug 2025  \\
		\hline
		\textbf{\texttt{gpt-oss-120b}} & Jun 2024 & Aug 2025 \\
		\hline
		\texttt{gemma-3-1b-it} & Aug 2024 & Mar 2025 \\
		\hline
		\texttt{gemma-3-4b-it} & Aug 2024 & Mar 2025 \\
		\hline
		\texttt{gemma-3-12b-it} & Aug 2024 & Mar 2025 \\
		\hline
		\texttt{gemma-3-27b-it} & Aug 2024 & Mar 2025  \\
		\hline
		\texttt{Llama-3.2-3B-Instruct} & Dec 2023 & Sep 2024 \\
		\hline
		\texttt{Llama-3.1-8B-Instruct} & Dec 2023 & Jul 2024 \\
		\hline
		\texttt{Llama-3.3-70B-Instruct} & Dec 2023 & Dec 2024 \\
		\hline
		\texttt{Llama-4-Maverick-Instruct-FP8} &  Aug 2024 & Apr 2025  \\
		\hline
		\texttt{phi-4} & June 2024 & Dec 2024\\
		\hline
		\texttt{Qwen3-1.7B} &  & Apr 2025 \\
		\hline
		\texttt{Qwen3-4B-Instruct-2507} &  & Aug 2025 \\
		\hline
		\texttt{Qwen2.5-7B-Instruct} &  & Sep 2024  \\  
		\hline
		\texttt{Qwen3-30B-A3B-Instruct-2507} &  & Jul 2025 \\
		\hline
		\textbf{\texttt{Qwen3-235B-A22B-Instruct-2507}} &  & Jul 2025  \\
		\bottomrule
	\end{tabular}
\end{table}

\subsection{Question Rewriting Procedure}

The question rewriting and expansion was approached as a pre-processing step before benchmarking the evaluation models against the modified questions. 
We evaluated three different LMs for question rewriting: \texttt{gpt-oss-20b} (small), \texttt{gpt-oss-120b} (medium), and \texttt{Qwen3-235B-A22B-Instruct-2507} (large).
For each question in each dataset the rewriting LM was provided the original question, correct answer, and grounding context during prompting (\Cref{apx:rewrite-prompt}) to rewrite the question to disambiguate what was being asked.
In addition to rewriting the question the rewriting model must generate the correct answer to the rewritten question, which serves as a validation check against topic drift as its later evaluated for semantic similarity to the original question. 
Note, the rewritten answer is never used. It only serves as a validation check.
During benchmark evaluation human curated answer is always the ``correct'' answer.
This results in three rewritten variants of every question, one for each of the rewriting models (bold models in \Cref{tab:model_cutoff}).

We filter the rewritten questions to ensure the that all final benchmark questions have the acceptable meta-properties. 
An LM-Judge (\texttt{gpt-oss-120b} prompt in \Cref{apx:question_validation}) is used to extract the following properties: reformatted question similarity to the original question, reformatted answer similarity to the original answer, reformatted question ``giveaway'' score, and original question ``giveaway'' score.
The reformatted question and answer similarity scores are used to discard any questions which are too dissimilar to the original.
The giveaway scores rate how much the answer was given away (or contained within) the question.
Any reformatted questions with giveaway scores higher than the original are discarded. 
This removes rewritten questions that are easier from the LM-judge's perspective, producing a new benchmark which is strictly more difficult, with disambiguated questions. 

\subsection{Evaluation methodology}

Evaluating the accuracy improvement provided by question rewriting and expansion 
requires benchmarking each model under evaluation for three configurations:
	\begin{packed_enum} 
		\item Original Question: the normal benchmarking use case, 
		\item Original Question with Context Prepended: the RAG retrieval use case, and 
		\item Rewritten Question: the improvement imparted by rewriting.
	\end{packed_enum}
Note that the ``Context'' could either contain the answer, or be Answer-Free.

To quantify the accuracy improvement, we compute the following average per-model and per-dataset differences:
\begin{packed_enum} 
	\item (Rewritten Question - Original Question): the direct rewrite improvement, 
	\item (Rewritten Question - Original Question with Context Prepended): the rewrite improvement compared to RAG retrieval
\end{packed_enum}

\noindent This comparison between the rewritten question and the original question with the context prepended enables evaluation of whether the rewrite outperforms benchmarking models with a RAG database which contains only background information.
However, if the context contains the answer, one would expect any benchmarking with the context included to yield very high accuracy.
To mitigate this we developed a version of each grounding context that is ``Answer-Free''. 
\texttt{AFC} was developed using a single LM (\texttt{gpt-oss-120b}) to rewrite the context (see \Cref{apx:afc} for the prompt).
The results section covers the accuracy improvement using the original context (the less interesting use case) as well as the Answer-Free Context (\texttt{AFC}).
\texttt{AFC} replicates the use case where a RAG system surfaces background information, but not the actual answer to the user's question. 
Our results demonstrate accuracy improvement in benchmark performance when question rewriting is performed using this Answer-Free Context. 
This improvement effect is larger than just prepending the \texttt{AFC}.

\begin{figure*}[t!b]
	\vspace{-2.5em} 
	\captionsetup[subfigure]{labelformat=empty}
	\centering
	\subfloat{\includegraphics[width=0.45\textwidth]{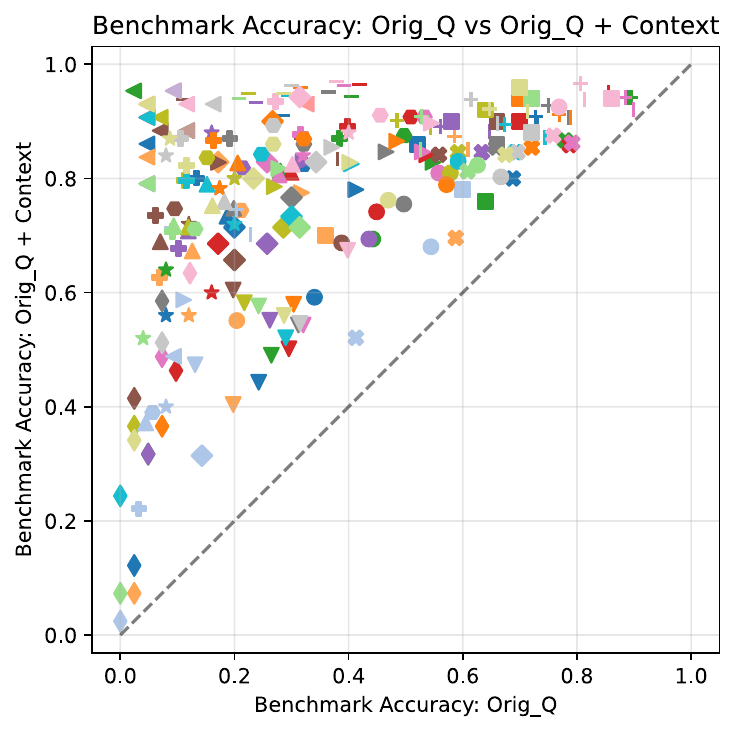}}
	\subfloat{\includegraphics[width=0.45\textwidth]{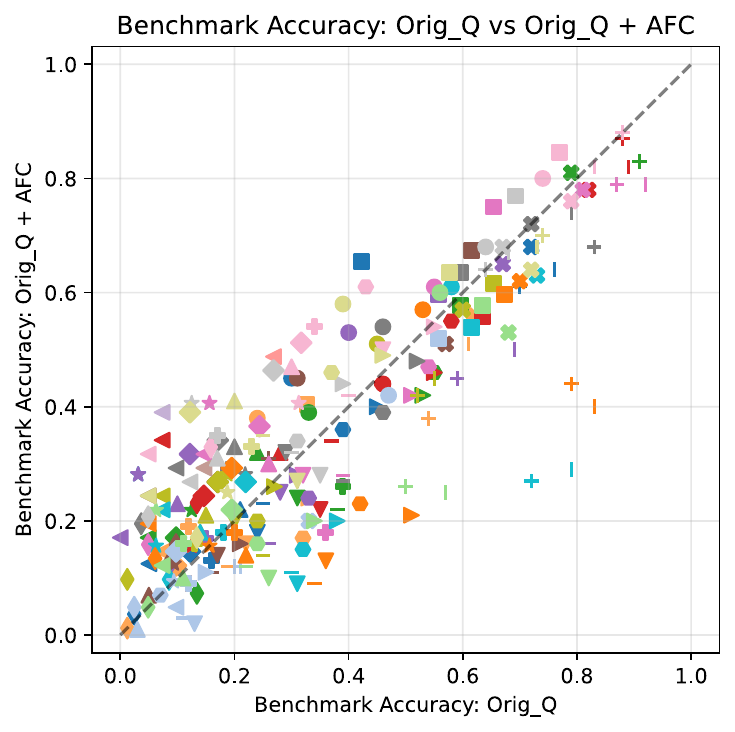}} \\
	\subfloat{\includegraphics[width=0.45\textwidth]{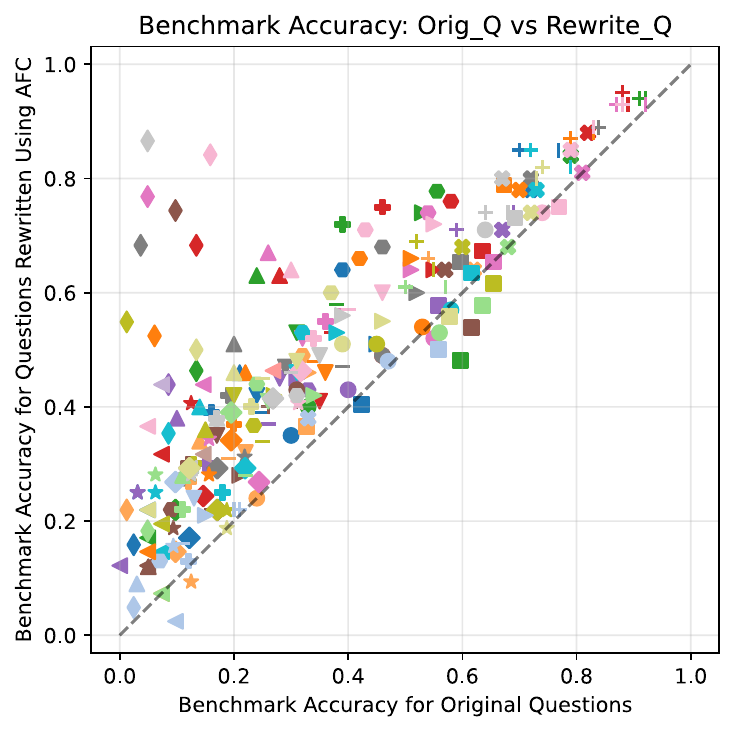}}
	\subfloat{\includegraphics[width=0.45\textwidth]{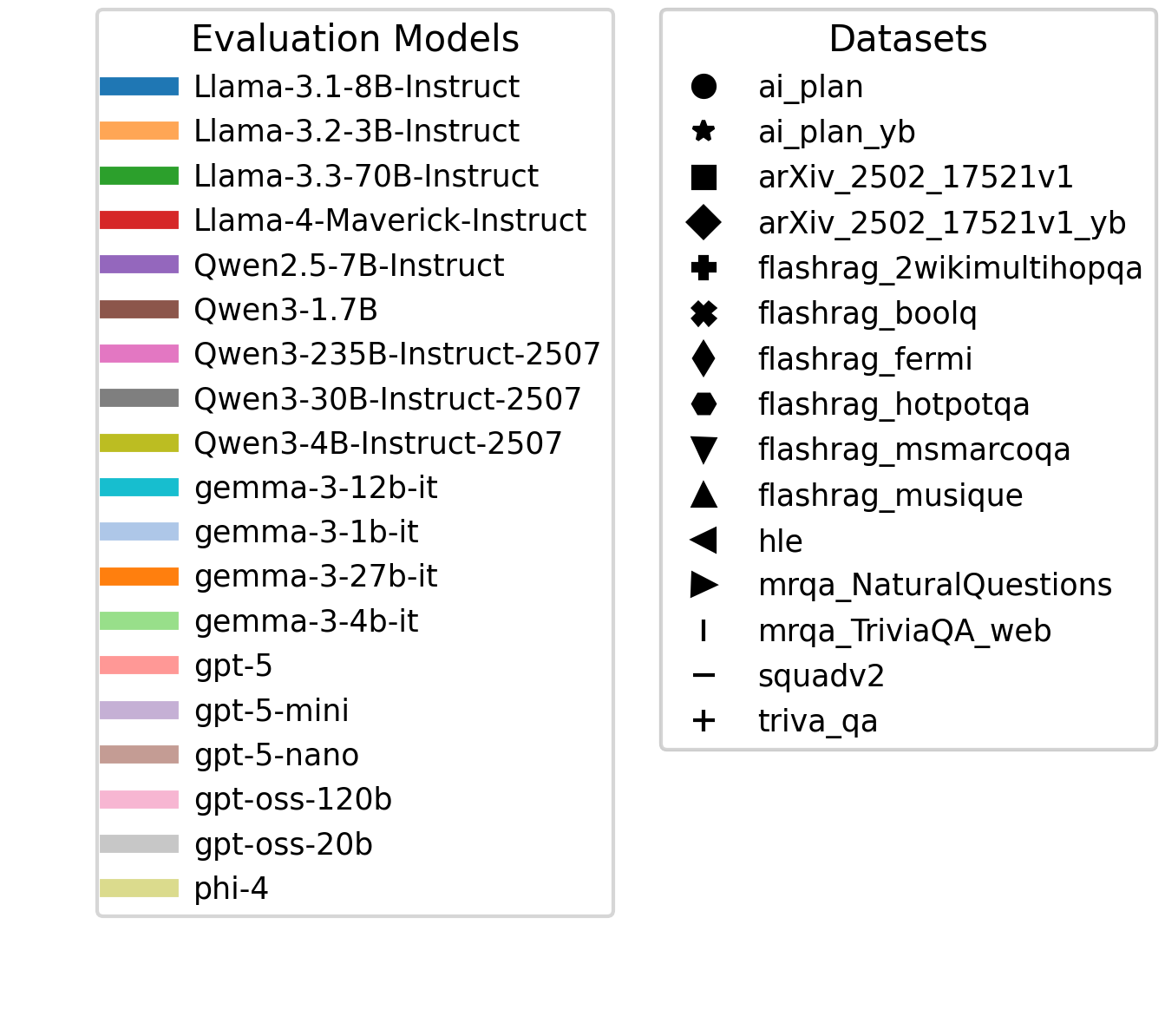}}
	\caption{Performance of various LMs with and without answer-containing context and answer-free context. The \textit{x}-axis shows the original question benchmark accuracy while the \textit{y}-axis shows the rewritten question benchmark accuracy; distance above the line conveys improvement. Colors represent the different models tested (\cref{tab:model_cutoff}) and shapes represent the dataset (\cref{tab:dataset_publication_cutoff}). (top left) \textit{Answers are present in the context \textbf{(baseline)}}: Benchmark performance significantly improves with the addition of context that contains the answer. (top right) \textit{Answers not present in the context}: Simply providing relevant answer free context without question rewriting does not improve performance. (bottom left) \textit{Questions rewritten using answer-free context \textbf{(our approach)}}: Improvement in accuracy caused by rewriting the question only using \texttt{AFC} where during benchmarking only the rewritten question is presented to the LM (\texttt{AFC} is withheld).}
	\label{fig:all-scatterplots}
\end{figure*}

To test one hypothesis for why this effect shows up, we leverage an embedding model (\texttt{e5-mistral-7b-instruct}) to understand the change in cosine distance between the original question and the context compared to the cosine distance between the rewritten question and the context.
We demonstrate that improvements in benchmark accuracy correlate with reductions in the cosine distance between the question and context.
In other words, the rewritten question appears to place the model closer to the right `frame of mind' to answer the question, as the cosine distance between the question and the context is smaller after the rewrite.
Assuming the answer-containing context points in a certain direction within the model's embedding space, rewriting the question moves the original question towards that correct answer direction.
Regardless of whether the original or rewritten question is given to the LM under test, the correct answer is always defined by the benchmark dataset and is never modified by the rewriting process.

\section{Results}



While the improved \texttt{gpt-5-mini} performance on HLE-subset (\Cref{fig:hle-uplift}) is already an informative result, in this section we investigate the impact of query disambiguation via Answer-Free Context for all models (\Cref{tab:model_cutoff}) and datasets (\Cref{tab:dataset_publication_cutoff}) across three dimensions:
\begin{enumerate*}[label=(\alph*)]
	\item the validity of our Answer-Free Context construction,
	\item the accuracy improvement provided by rewriting compared to standard RAG baselines, and 
	\item the underlying mechanisms driving these performance gains.
\end{enumerate*}

\subsection{Validation of Answer-Free Context} 

To ensure that performance gains result from query disambiguation rather than information leakage, we first verify that our Answer-Free Context does not inadvertently contain the answer to the benchmark question. 
\Cref{fig:all-scatterplots} (top left) illustrates performance when the correct answer \textit{is present} in the context. As expected, prepending answers yields a significant improvement in accuracy. 
This confirms that models successfully utilize answer information when it is present. 
In contrast, \Cref{fig:all-scatterplots} (top right) plots model performance when \texttt{Answer-Free Context} is prepended to the original question. 
The absence of consistent performance improvement shows that the AFC creation process successfully removed the direct answer.  
Consequently, any accuracy improvement in subsequent experiments can be attributed to query disambiguation rather than rote answer extraction.

\subsection{Accuracy improvement from Query Rewriting} 
We compare the performance of the Rewritten Question (\texttt{Rewrite\_Q}) against two baselines: the Original Question (\texttt{Orig\_Q}) and the Original Question with AFC prepended (\texttt{Orig\_Q+AFC}).
Unless otherwise stated the \texttt{Rewrite\_Q} condition never includes \texttt{AFC}, it is only the rewritten question.

\paragraph{General improvement} 
As shown in \Cref{fig:hle-uplift}, applying the rewrite strategy to the HLE-subset yields substantial gains, improving \texttt{gpt-5-mini} accuracy from 13.9\% (\texttt{Orig\_Q}) to 37.2\% (\texttt{Rewrite\_Q}). 
This trend generalizes across datasets.
On average, asking the \texttt{Rewrite\_Q} questions provides an accuracy improvement of 13.03\% compared to \texttt{Orig\_Q} questions when neither have the \texttt{AFC} included with the question.
The distribution of benchmark accuracies across all models and datasets is available in \Cref{apx:rewrite_impact}.

\Cref{fig:all-scatterplots} (bottom left) further breaks this accuracy improvement down by dataset, showing that most (model,dataset) combinations fall above the identity line ($y=x$), confirming that rewriting rarely degrades performance relative to the original query (at the benchmark level), and never degrades performance for any dataset pre-model knowledge cutoff.

Additionally, some datasets like \texttt{flashrag\_fermi} demonstrate significant improvement stemming from disambiguation.

\begin{figure}[h!]
	\vspace{-0.5em}
	\centering
	\includegraphics[width=0.9\columnwidth]{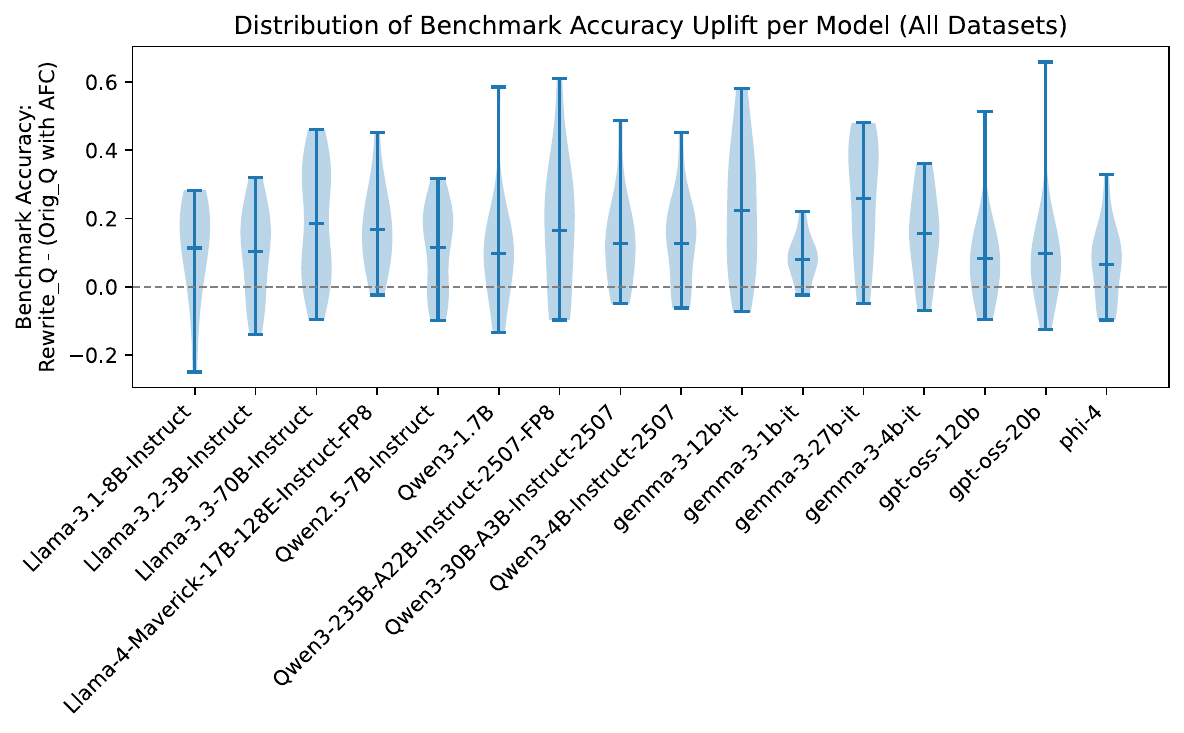}
	\vspace{-0.8em}
	\caption{Per-dataset per-model difference in benchmark accuracy between the rewritten questions and the original questions with associated answer-free context. The violin plot distribution highlights the range of accuracy deltas over all datasets for each model evaluated. Benchmark accuracy improved by an average of 0.1346.}
	\label{fig:r_minus_q_afc_giveaway}
	\vspace{-1em}   
\end{figure}

\paragraph{Comparison to RAG Baselines}
Considering that rewritten questions typically contain more detailed information drawn from context, comparing them directly against the original is not a fair comparison. 
A stricter evaluation compares results on the rewritten question against a standard RAG approach where the model receives the original question with the context prepended ($\texttt{Orig\_Q+AFC}$).
This isolates the value of the rewrite versus simply having access to the \texttt{AFC} information. 
We note that the question rewriting process only uses the answer-free context, so the rewritten question and the original question with AFC are on even footing information-wise.
\Cref{fig:r_minus_q_afc_giveaway} displays the distribution of ($\texttt{Rewrite\_Q} - \texttt{Orig\_Q+AFC}$).
Note that during evaluation \texttt{Rewrite\_Q} does not include \texttt{AFC}, it only presents the rewritten question to the model; whereas \texttt{Orig\_Q+AFC} asks the original question after the \texttt{AFC}.

For the majority of questions across all models and datasets, the rewrite strategy outperforms the RAG baseline. 
Each violin plot (per-model) is a distribution of benchmark accuracy deltas.
In this case, the model under evaluation (\texttt{Orig\_Q+AFC}) has access to all information the rewriting model did. 
This demonstrates that rewriting the questions in a separate rewrite-then-answer paradigm produces an accuracy improvement that is not achieved by prepending the same information into the models' context during evaluation.
\FloatBarrier

\begin{figure}[h!]
	\vspace{-0.5em}
	\centering
	\includegraphics[width=0.9\columnwidth]{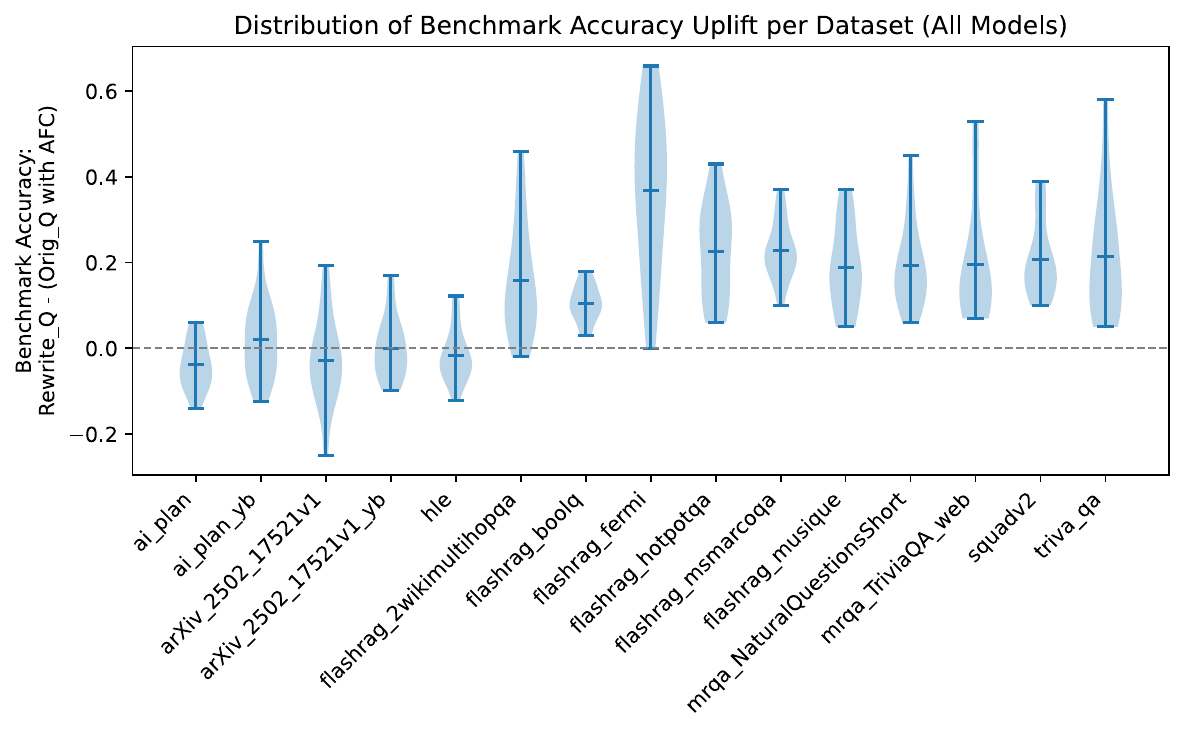}
	\vspace{-0.8em}
	\caption{Per-dataset difference in benchmark accuracy between the rewritten questions without \texttt{AFC} and the original questions with \texttt{AFC}. The violin plot distribution highlights the range of accuracy deltas over all models for each dataset evaluated. Benchmark accuracy improved by an average of 0.1346.}
	\label{fig:r_minus_q_afc_giveaway_dataset}
	\vspace{-0.5em}   
\end{figure}

\paragraph{The Reasoning Gap}  
A notable exception is observed in benchmarks that focus less on fact recall (are more reasoning intensive), specifically the HLE-subset and generative benchmarks. 
As detailed in \Cref{fig:r_minus_q_afc_giveaway_dataset}, these datasets show no consistent improvement when comparing \texttt{Rewrite\_Q} to \texttt{Orig\_Q+AFC}. 
While rewriting improves on the isolated \texttt{Orig\_Q} query, more complex reasoning tasks benefit more from having the full raw context available in the window during inference than from a condensed, rewritten query. 

In \Cref{fig:r_minus_q_afc_giveaway_dataset}, the left-most five datasets are from after the knowledge cutoff (HLE,  ai\_plan \& arXiv\_2502\_17521v1). 
Additionally, due to a paucity of datasets released since the beginning of the year, all but HLE-subset were generatively created using the methods of either~\citet{majurski2025generative} or~\citet{shashidhar2025yourbench}.
Those benchmarks tend to have more complex questions that are less fact-based than the extractive QA datasets which make up the majority.

Unfortunately, the datasets available for this study have two conflated effects.
The more reasoning-intensive benchmarks (generative and HLE-subset) are also the datasets that are post knowledge cutoff for all models (see \Cref{tab:dataset_publication_cutoff} and \Cref{tab:model_cutoff}. 
It is possible either that the lack of improvement from rewriting stems from the questions not being subject to training data contamination, or from the fact that the questions require more reasoning and less recall.
This indicates a divergence in optimal strategies, with factual disambiguation favoring a rewrite and more complex queries favoring raw context inclusion.

\begin{figure}[h!]
	\vspace{-0.5em}
	\centering
	\includegraphics[width=0.9\columnwidth]{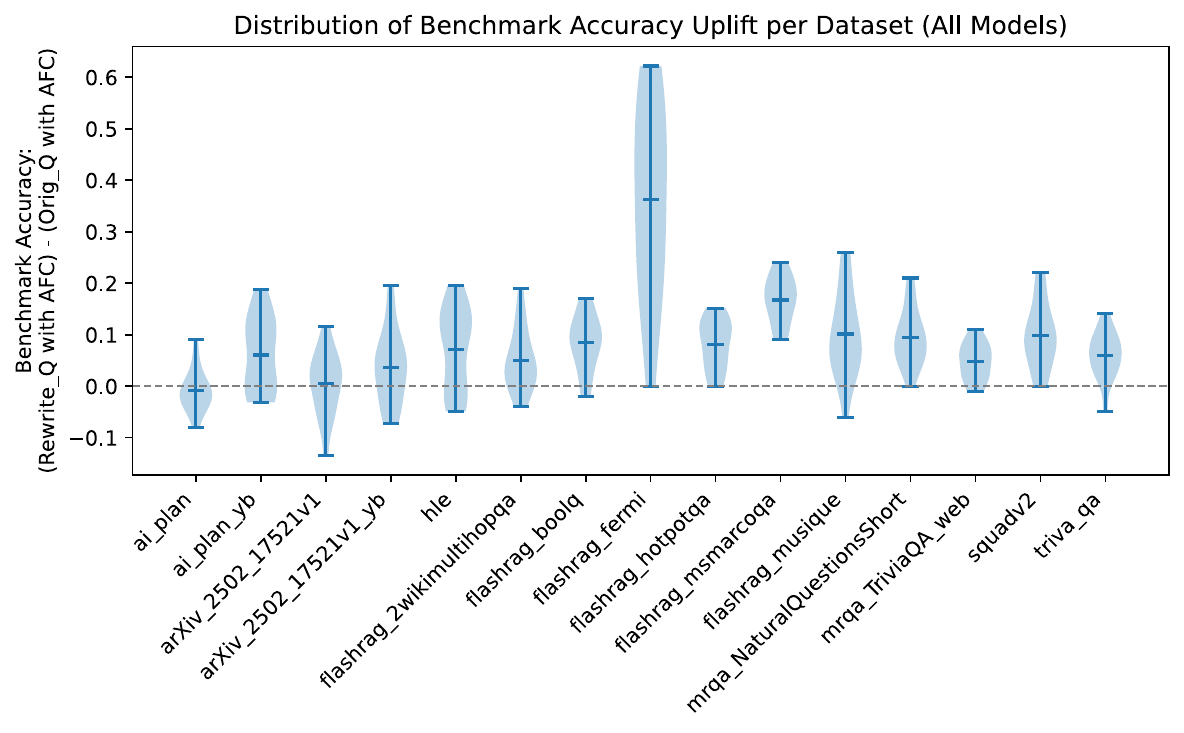}
	\vspace{-0.8em}
	\caption{Per-dataset improvement in benchmark accuracy from \texttt{Rewrite\_Q+AFC} compared to the \texttt{Orig\_Q+AFC} during benchmark evaluation. Benchmark accuracy improved by an average of 0.0875.}
	\label{fig:rafc_giveaway_minus_qafc_giveaway_dataset}
	\vspace{-0.5em}   
\end{figure}

Combining both strategies limits the potential downsides to rewriting shown in \Cref{fig:r_minus_q_afc_giveaway_dataset} where benchmark performance might be reduced slightly from the rewrite (compared to original questions with \texttt{AFC}). 
\Cref{fig:rafc_giveaway_minus_qafc_giveaway_dataset} demonstrates that \texttt{Rewrite\_Q+AFC} limits the potential accuracy improvement (average improvement of 0.0875 instead of 0.1346) but reduces the number of datasets which show reductions in accuracy.
This may indicate that disambiguation and context inclusion are complementary.

\subsection{Alignment and Task Separation} 
\label{sec:task_seperation}



\begin{figure}[h!]
	\vspace{-0.5em}
	\centering
	\includegraphics[width=0.95\columnwidth]{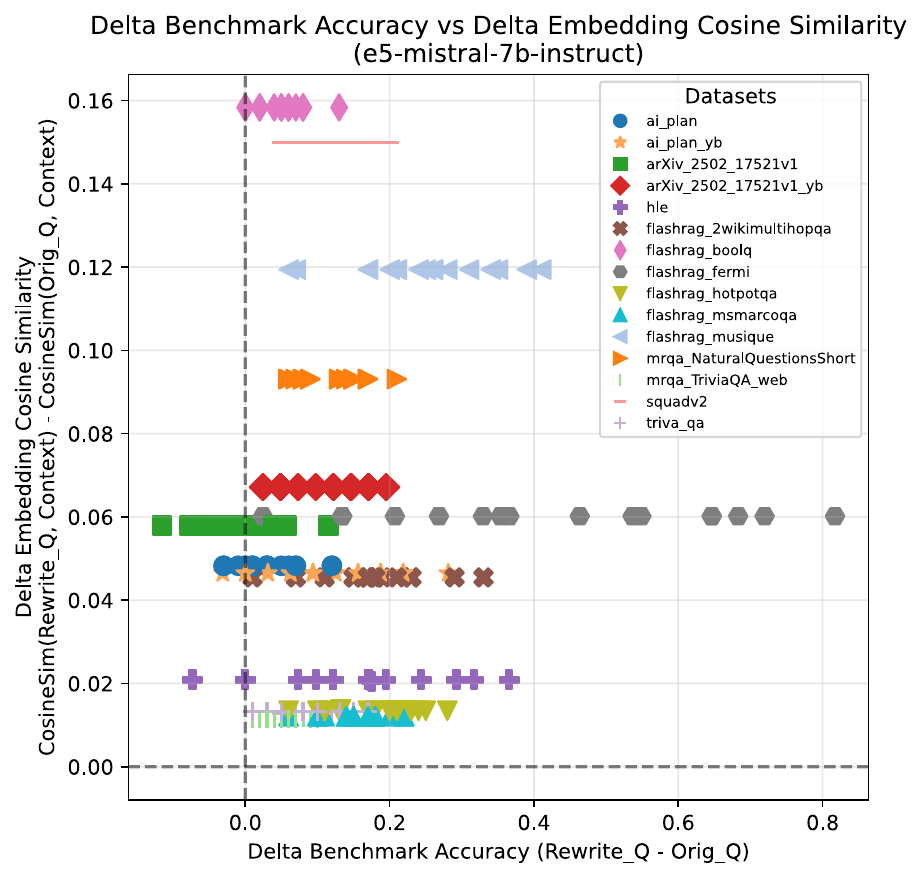}
	\vspace{-0.8em}
	\caption{Accuracy improvement on the \textit{x}-axis \textit{vs.} improvement in cosine similarity between the question and context for rewritten questions on the \textit{y}-axis. \textit{x}-axis: delta benchmark accuracy \texttt{Rewrite\_Q $-$ Orig\_Q}; larger values indicate that the rewritten question improved benchmark accuracy. \textit{y}-axis: increase in cosine similarity between the question and context due to rewriting; larger values indicate better alignment between \texttt{Rewrite\_Q} and \texttt{AFC} compared to \texttt{Orig\_Q} and \texttt{AFC}. The fact that most points are in the upper right quadrant demonstrates that improvements in embedding alignment between the question and grounding context correlate with improved benchmark accuracy.}
	\label{fig:acc_vs_embedding_gpt120b_e5-mistral-7b-instruct}
	\vspace{-0.5em}   
\end{figure}

To understand the drivers behind the accuracy improvements, we analyze the semantic alignment between queries (questions) and their grounding contexts. 
\Cref{fig:acc_vs_embedding_gpt120b_e5-mistral-7b-instruct} plots the change in downstream benchmark accuracy (\textit{x}-axis) against the change in query-context cosine similarity (\textit{y}-axis) when changing from the original query (\texttt{Orig\_Q}) to the rewritten query (\texttt{Rewrite\_Q}), measured using the \texttt{e5-mistral-7b-instruct} embedding model. 
We observe a strong positive relationship: Rewritten questions systematically exhibit higher cosine similarity to the context than the original questions. 
This effect is highly consistent, with $91\%$ of the dataset points (222 out of 243) falling into the upper-right quadrant ($x>0, y>0$). 
This clustering demonstrates that queries rewritten to have tighter semantic alignment with the context reliably produce higher benchmark accuracy.

It is worth noting that every model and dataset combination drawn from before the knowledge cutoff (where training data contamination might be in effect) appear in the upper right quadrant.
However, for some models in the post-knowledge cutoff datasets, there is a drop in benchmark accuracy, as evidenced by those points in the upper left quadrant. 
Each dataset has a distribution of accuracy improvements across the evaluated models, which is why the datasets in \Cref{fig:acc_vs_embedding_gpt120b_e5-mistral-7b-instruct} form horizontal bands.
This suggests the rewriting process effectively aligns the query with the latent space of the relevant information from the context.

\begin{figure}[h!]
	\vspace{-0.5em}
	\centering
	\includegraphics[width=0.9\columnwidth]{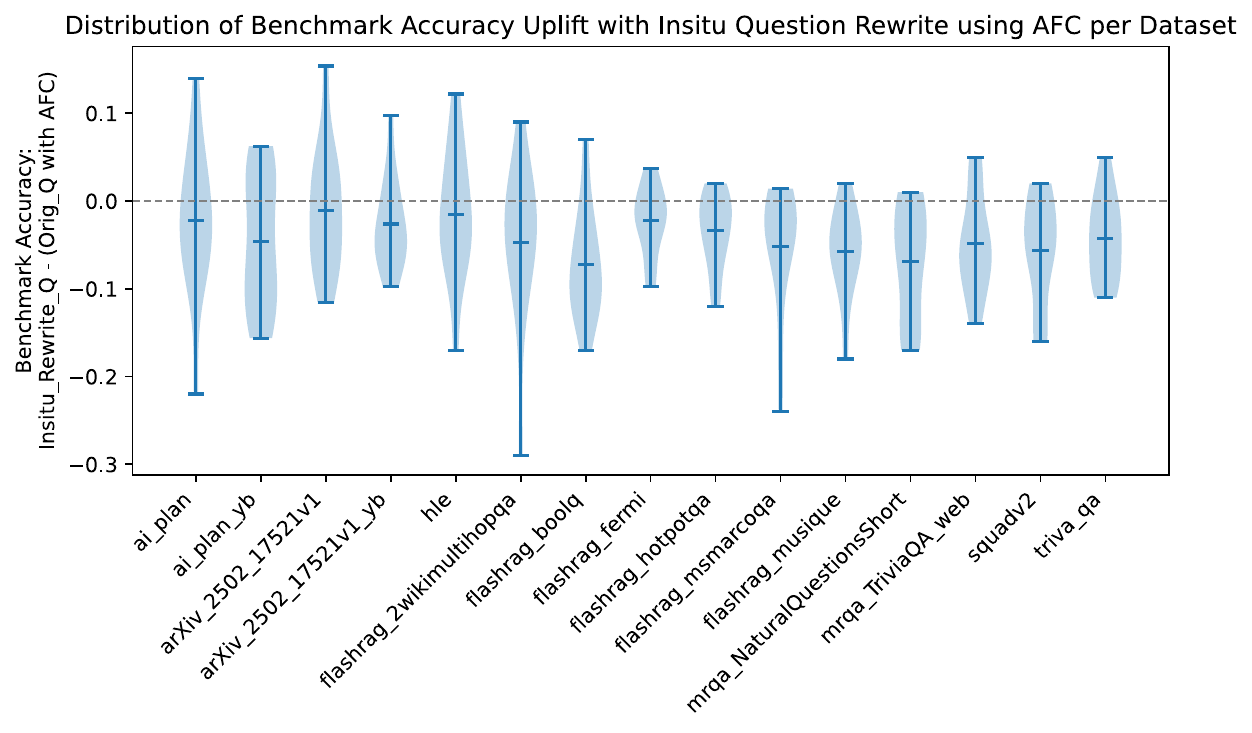}
	\vspace{-0.8em}
	\caption{Per-dataset improvement in benchmark accuracy from performing an \textit{in situ} rewrite of the question using answer-free context during benchmark evaluation. This combines the rewrite-then-answer method into a single operation. The prior accuracy improvement disappears, highlighting the impact of task separation between the rewrite and answer phases.}
	\label{fig:insitu_rafc_minus_q_afc_giveaway_dataset}
\end{figure}

Finally, we test whether this rewriting must be an explicit preprocessing step or if it can be induced via Chain-of-Thought (CoT). 
We implemented an ``In Situ'' baseline where the model is prompted to rewrite the question internally using identical prompts and \texttt{AFC} before answering. 
\Cref{fig:insitu_rafc_minus_q_afc_giveaway_dataset} shows that the accuracy improvement disappears under this paradigm. 
This holds regardless of whether the model supports \texttt{<thinking>} tags. 
CoT rewriting was evaluated using the same prompt, lightly modified to have the LM first rewrite for disambiguation before answering the rewritten question. 
This result highlights the necessity of task separation: the `cognitive load' or context window dynamics of rewriting and answering in a single pass negates the benefits of disambiguation.

\subsection{Limitations}

Our evaluations into the impact of question rewriting rely heavily on extractive QA datasets which have pairs of question and context.
The exception is HLE-subset, where the questions are \textit{post facto} grounded using internet search by domain experts~\citep{HleFutureHouse}.
Thus, many of the fact-based extractive QA dataset questions are easily answerable when the LM is presented the answer-containing context. 
This limitation---which can be summarized as the \texttt{AFC} available with the datasets originally contained the answer---is an inevitable consequence of the availability of public datasets containing questions paired with grounding context.

We use LM-as-a-judge throughout, following ~\cite{majurski2026grounding}, in which LM-judge approaches were human validated. Nonetheless, future work in which one or more judge components are replaced with (possibly crowdsourced) human evaluation would strengthen these findings further.

\section{Conclusion}

This work demonstrates a novel utilization of the information-dynamic context retrieval systems surface for a query.
RAG utility extends beyond a binary success or failure of surfacing the direct answer; we introduce and validate query disambiguation leveraging answer-free context information. 
This demonstrates that surfaced background information can enhance LM performance even when it does not contain the answer. 
By using \texttt{AFC} to rewrite and disambiguate user queries prior to inference, we observed substantial accuracy improvement across multiple benchmarks, most notably doubling the performance of \texttt{gpt-5-mini} on a subset of Humanity's Last Exam (HLE).

Our analysis produces three insights for the design of future dynamic context grounded systems: 
(1) The mechanism of rewrite improvement is measurable and predictable, as accuracy gains correlate with increased semantic alignment (cosine similarity) between the rewritten query and the grounding context. 
(2) We identify two behaviors in response to query rewriting: Primarily factual queries benefit maximally from rewriting alone; more complex questions (such as HLE) achieve peak performance when the model is provided with both the disambiguated query and the raw context. 
(3) We establish the necessity of task separation. The inability to replicate rewriting-driven performance improvement during an \textit{in situ} Chain-of-Thought experiment implies that query refinement and answer generation compete for cognitive resources or context attention.
This holds true for models both with and without formalized ``reasoning'' capability. 

This work suggests treating dynamic context construction methodologies as not purely as evidence collectors, but as a collaborative partner in query formulation.
Aligning the user's query to the document corpus and clarifying the requested information allows LMs to more precisely and correctly provide the information a user actually wants.

\subsection{Future work}
This study only used single-shot query rewriting.
Accuracy improvements are likely possible by extending this to a best-of-N approach that attempts to disambiguate the user's questions in multiple directions based on surfaced evidence, before asking the user which direction actually aligns with their intent. 
Another improvement would be a single multi-turn process which surfaces evidence, rewrites the query, and then performs another search of the corpus with the clarified query.
Additional work is required to characterize the additional inference time costs to this approach (how many additional tokens are being spent and RAG database calls made to perform query disambiguation) or how small an LM can be used for this process while still maintaining result fidelity. This methodology provides another test time parameter that can be modified depending on the requirements of the situation.

{\small
	\bibliography{references}
}

\section*{Appendices}
\begin{appendices}
	\crefalias{section}{appendix}
	\appendix
	
\section{Impact of Rewrite on Benchmarks}
\label{apx:rewrite_impact}

\begin{figure}[h!]
	\vspace{-0.5em}
	\centering
	\includegraphics[width=0.9\columnwidth]{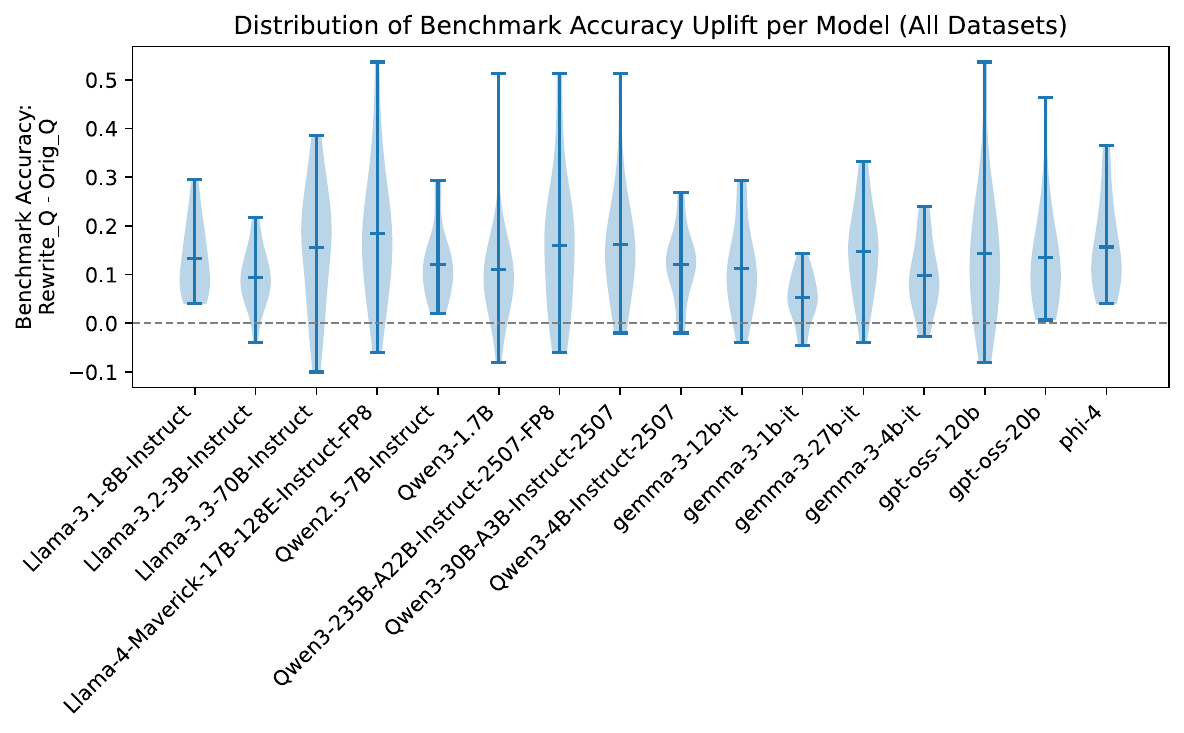}
	\vspace{-0.8em}
	\caption{Per-dataset per-model difference in benchmark accuracy between the rewritten question and the original question. The violin plot distribution highlights the range of accuracy deltas over all datasets for each model evaluated. Benchmark accuracy improved by an average of 0.1303.}
	\label{fig:r_minus_q_apx}
\end{figure}

\Cref{fig:r_minus_q_apx} presents the distribution of accuracy improvement measured as ($\texttt{Rewrite\_Q} - \texttt{Orig\_Q}$) per-evaluation model across all datasets. 
This demonstrates question rewriting broadly improves benchmark accuracy. 
\FloatBarrier

\begin{figure}[h!]
	\vspace{-0.5em}
	\centering
	\includegraphics[width=0.9\columnwidth]{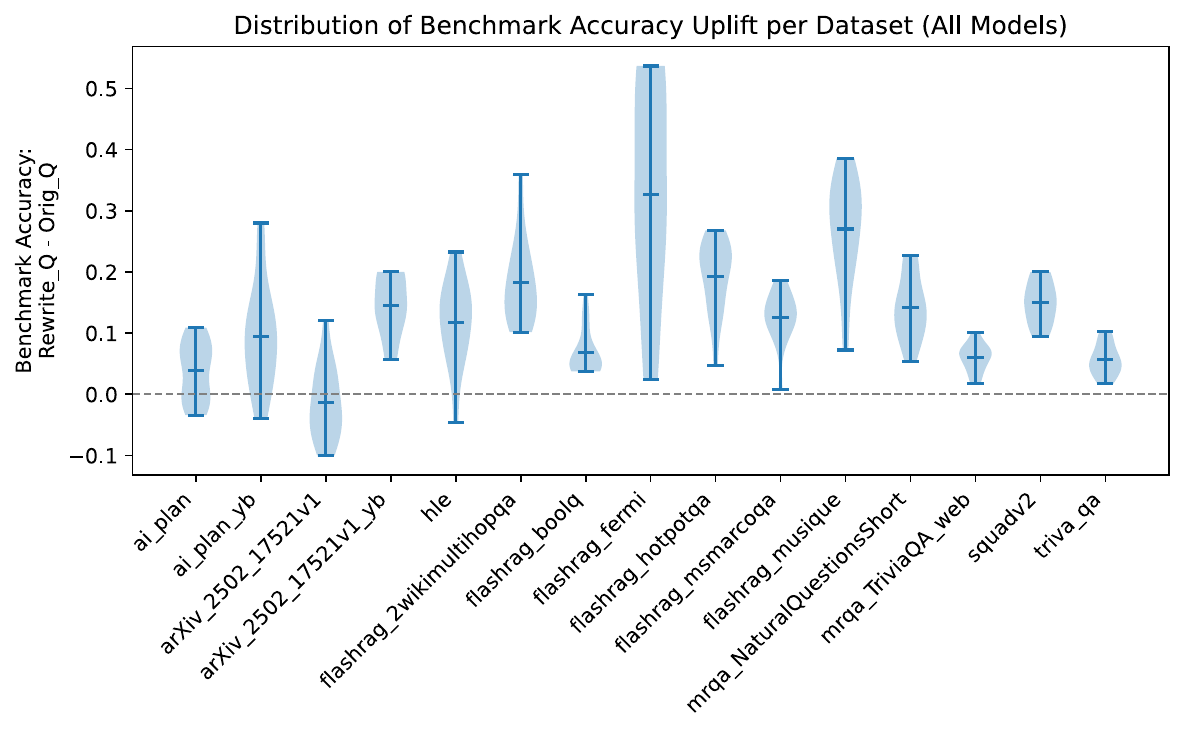}
	\vspace{-0.8em}
	\caption{Per-dataset per-model difference in benchmark accuracy between the rewritten question and the original question. The violin plot distribution highlights the range of accuracy deltas over all datasets for each model evaluated. Benchmark accuracy improved by an average of 0.1303.}
	\label{fig:r_minus_q_dataset_apx}
\end{figure}

\Cref{fig:r_minus_q_dataset_apx} presents the distribution of accuracy improvement measured as ($\texttt{Rewrite\_Q} - \texttt{Orig\_Q}$) per-dataset instead of per-model like \Cref{fig:r_minus_q_apx}.
\FloatBarrier

\begin{figure}[h!]
	\vspace{-0.5em}
	\centering
	\includegraphics[width=0.9\columnwidth]{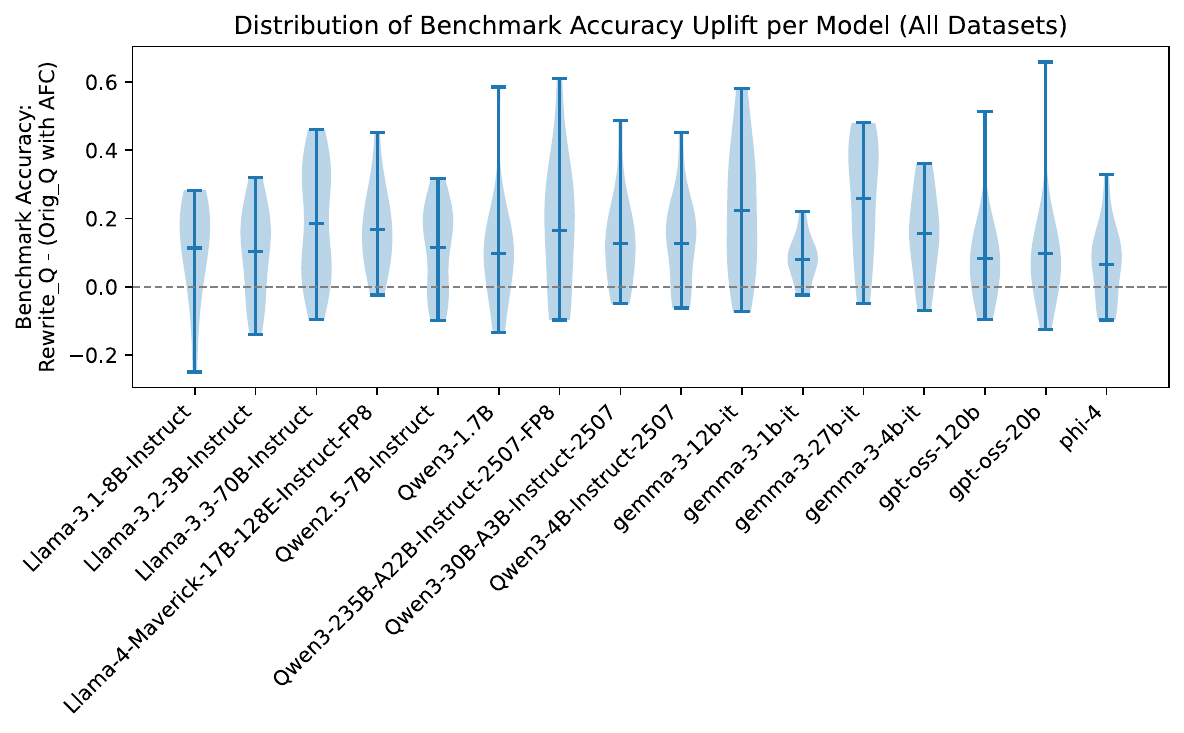}
	\vspace{-0.8em}
	\caption{Per-model difference in benchmark accuracy between the rewritten questions and the original questions with associated answer-free context. The violin plot distribution highlights the range of accuracy deltas over all datasets for each model evaluated. Benchmark accuracy improved by an average of 0.1346.}
	\label{fig:r_minus_q_afc_giveaway_apx}
\end{figure}

\Cref{fig:r_minus_q_afc_giveaway_apx} presents the distribution of accuracy improvement measured as ($\texttt{Rewrite\_Q} - \texttt{Orig\_Q+AFC}$) per-evaluation model across all datasets. 
\FloatBarrier

\begin{figure}[h!]
	\vspace{-0.5em}
	\centering
	\includegraphics[width=0.9\columnwidth]{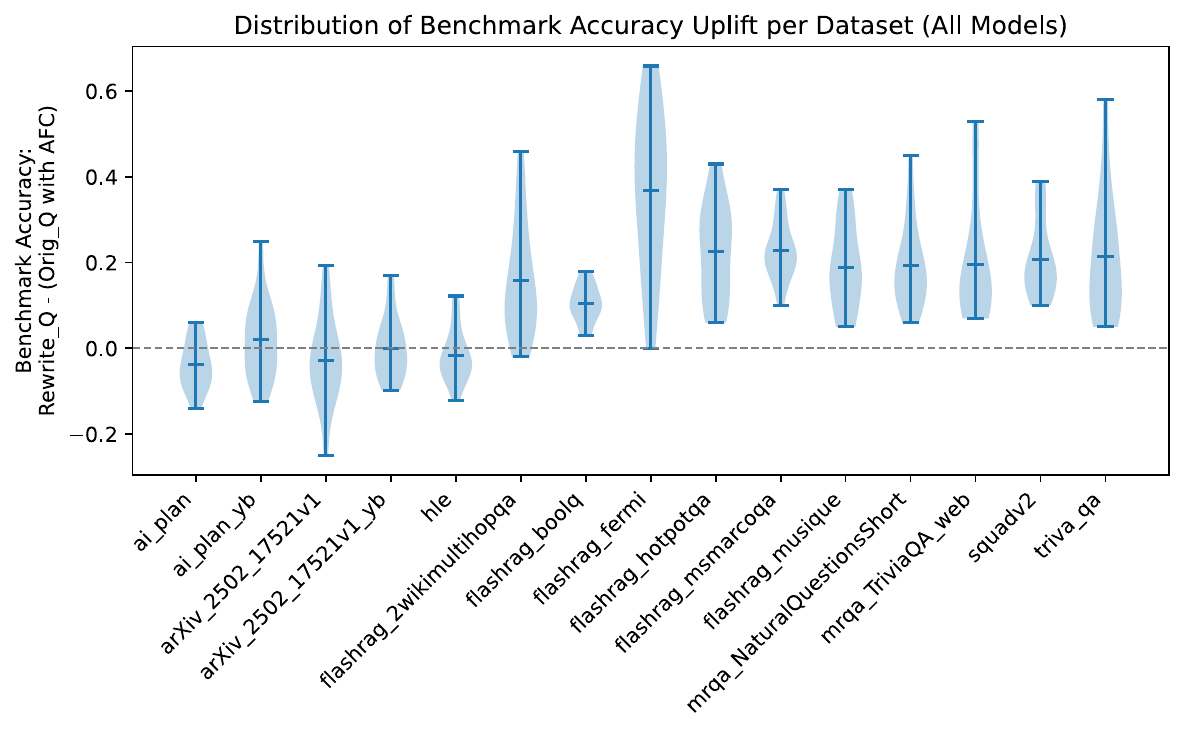}
	\vspace{-0.8em}
	\caption{Per-dataset per-model difference in benchmark accuracy between the rewritten questions and the original questions with associated answer-free context. The violin plot distribution highlights the range of accuracy deltas over all models for each dataset evaluated. Benchmark accuracy improved by an average of 0.1346.}
	\label{fig:r_minus_q_afc_giveaway_dataset_apx}
\end{figure}

\Cref{fig:r_minus_q_afc_giveaway_dataset_apx} presents the same information as \Cref{fig:r_minus_q_afc_giveaway_apx}, but with each violin distribution per-dataset instead of per-model.
\FloatBarrier

\begin{figure}[h!]
	\vspace{-0.5em}
	\centering
	\includegraphics[width=0.9\columnwidth]{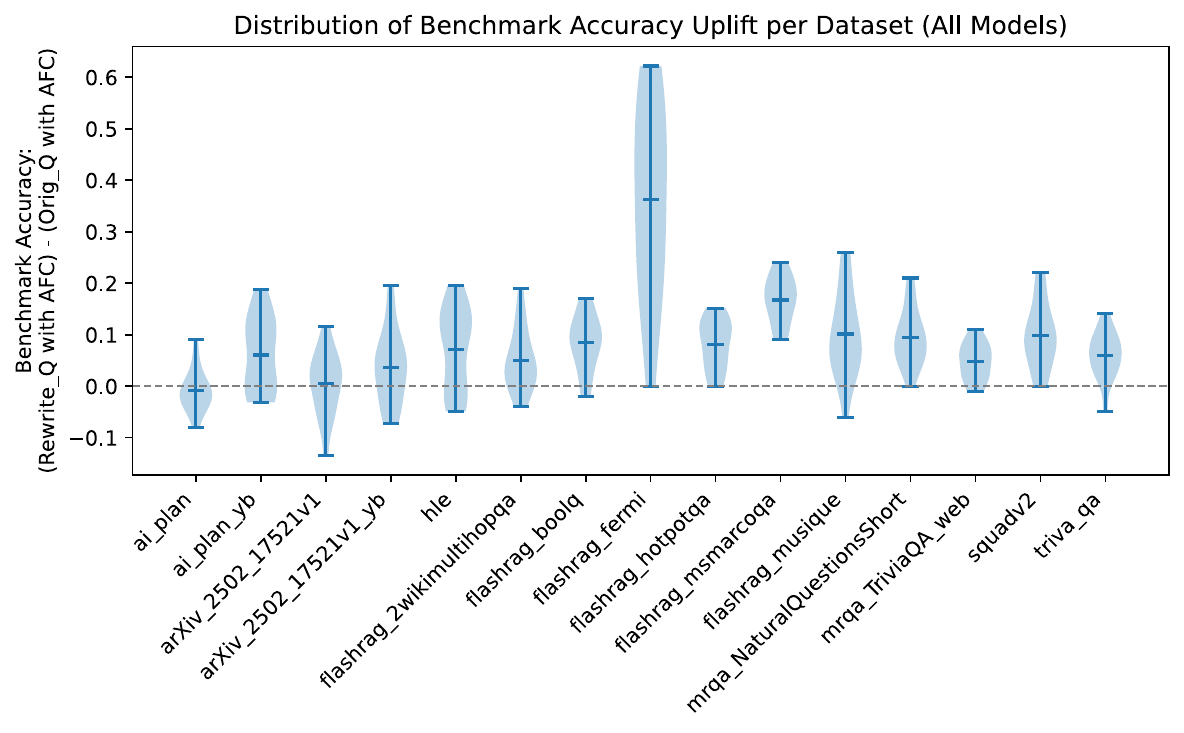}
	\vspace{-0.8em}
	\caption{Per-dataset improvement in benchmark accuracy from \texttt{Rewrite\_Q} with \texttt{AFC} compared to the \texttt{Orig\_Q} with \texttt{AFC} during benchmark evaluation. Benchmark accuracy improved by an average of 0.0875.}
	\label{fig:rafc_giveaway_minus_qafc_giveaway_dataset_apx}
\end{figure}

\Cref{fig:rafc_giveaway_minus_qafc_giveaway_dataset_apx} demonstrates the benchmark accuracy distribution per-dataset of \texttt{Rewrite\_Q+AFC - Orig\_Q+AFC}.
This combination limits the potential accuracy improvement but reduces the number of datasets which show no average improvement in accuracy.
This may indicate that disambiguation and context inclusion are complementary.
\FloatBarrier

\section{Question Rewriting Prompt}
\label{apx:rewrite-prompt}

{\scriptsize 
	\begin{lstlisting}[language=json]
		QUESTION_REFORMAT_PROMPT = """
		## Your Role
		
		You are an expert educational content creator specializing in editing and improving evaluation questions to determine the competency of domain experts based on the provided textual information. 
		
		## Input Structure
		
		Your input consists of:
		
		<question>
		[A question to be answered.]
		</question>
		
		<answer>
		[The correct answer to the question.]
		</answer>
		
		<context>
		[The text segment containing information relevant to the question.]
		</context>
		
		## Primary Objective
		
		Your goal is to reformat, rephrase, and rewrite the question according to the provided instructions. The rewritten question should be semantically equivalent to the original question, rewritten for clarity while preserving the same correct answer. This should only be accomplished by filling in background information and explicitly stating assumptions. You are creating a test/quiz question, so DO NOT include the answer information in the question, as that would be a giveaway which skews the results. NEVER include the answer or information which would give away the answer in the rewritten question.
		
		## Analysis Phase
		
		Conduct careful analysis within `<document_analysis>` tags, following these steps:
		
		1. **Thoughtful Content Examination**
		- Carefully analyze the given context, question, and answer; identifying central ideas, nuanced themes, and significant relationships within it.
		
		2. **Concept Exploration**
		- Consider implicit assumptions, subtle details, underlying theories, and potential applications of the provided information.
		
		3. **Intentional Question Planning**
		- Plan how the question can invite deeper understanding, meaningful reflection, or critical engagement, ensuring the question is purposeful.
		
		4. **Detailed Assumption Expansion**
		- Consider what knowledge the question is asking about, and what information and assumptions have been made when formatting the question. Your goal is to provide all the background information and explicitly state assumptions to enhance the clarity of the question.
		
		5. **Giving Away the Answer**
		- Plan how to avoid giving away the answer in the rewritten question. 
		- NEVER include the answer or information which would give away the answer in the rewritten question.
		
		### Documentation in Analysis:
		
		- Clearly document the rationale in the `<document_analysis>` tags, explaining your reasons for exclusion or inclusion decisions.
		- Clearly document what elements of the question need to be disambiguated. What steps need to be taken and what information needs to be include most clearly and concisely disambiguate the question. 
		- Clearly document what information needs to be avoided in the rewritten question to prevent giving away the answer. For example if the question asks about what year a person was born, the question should not include birthday in the biographical details.
		
		
		## Question Rewriting Guidelines
		
		### Encouraged Question Characteristics:
		
		- **Thoughtful Engagement**: Prioritize creating questions that inspire deeper thought and nuanced consideration.
		- **Deep Understanding and Insight**: Ensure that the question and answers require a deep understanding of the content by a professional domain expert.
		- **Self-contained Clarity**: Questions and answers should contain sufficient context, clearly understandable independently of external references.
		- **Brevity**: The rewritten question should be as short as is reasonable while still being clear, understandable, self-contained, and unambiguous.
		
		### Permitted Question Types:
		
		- Analytical
		- Application-based
		- Clarification
		- Counterfactual
		- Understanding
		- Conceptual
		- Factual
		- Open-ended
		- False-premise
		- Edge-case
		- Inference
		- Implication
		- Prediction
		
		(You do not need to use every question type, only those naturally fitting the content and instructions.)
		
		## Output Structure
		
		Present your final output strictly adhering the `<output_format>` tags.
		<output_format>
		Question: [ Question Text ]
		Explanation: [Brief explanation of why the answer is correct]
		Correct Answer: [Short answer]
		</output_format>
		
		## Output
		
		Begin by thoughtfully analyzing the provided context within `<document_analysis>` tags. Then present the resulting formatted question answer pair clearly within `<output_format>` tags.
		
		## Important Notes
		
		- NEVER modify the core element the question is asking about. The knowledge being evaluated shall not change. 
		- Question disambiguation and modification must be grounded in the `<context>`. 
		- Maintain clear, direct, and accurate citations/explanations drawn verbatim from the provided context.
		- Each "thought_process" should reflect careful consideration and reasoning behind your response.
		- When rewriting questions, NEVER include phrases like 'as per the text,' 'according to the document,' or any similar explicit references. Questions should inherently integrate content naturally and stand independently without explicit references to the source material. Make sure that the question is answerable by a domain expert **without the context paragraph**. 
		- Include all relevant context information in the question. Make the question as long and detailed as required so that the test taker can fully understand what is being asked.
		- NEVER include the answer in the rewritten question.
		- Ensure rigorous adherence to output formatting and generate a single `<output_format>` tag block.
		- Verify that the correct answer is in fact correct and the best version of that answer.
		- Verify that the question and answer are semantically equivalent to the original question and answer.
		
		
		
		<question>{question}</question>
		<answer>{answer}</answer>
		<context>{context}</context>
		"""
	\end{lstlisting}
}

\section{Answer-Free Context Creation}
\label{apx:afc}

{\scriptsize 
	\begin{lstlisting}[language=json]
		ANSWER_FREE_CONTEXT_PROMPT = """
		## Your Role
		
		You are an expert educational content creator specializing in editing and improving evaluation questions to determine the competency of domain experts based on the provided textual information. 
		
		## Input Structure
		
		Your input consists of:
		
		<question>
		[A question to be answered.]
		</question>
		
		<answer>
		[The correct answer to the question.]
		</answer>
		
		<context>
		[The text segment containing information relevant to the question.]
		</context>
		
		## Primary Objective
		
		Your goal is to reformat, rephrase, and rewrite the context information according to the provided instructions. The rewritten context should be minimally modified, and semantically equivalent to the original context. The rewrite should only remove the information which gives away the answer to the question. You are creating background material for a test/quiz question, so you need to COMPLETLELY remove the information which gives away the answer to the question from the context. NEVER include the answer or information which would give away the answer in the rewritten context.
		
		## Analysis Phase
		
		Conduct careful analysis within `<document_analysis>` tags, following these steps:
		
		1. **Thoughtful Content Examination**
		- Carefully analyze the given context, question, and answer; identifying central ideas, nuanced themes, and significant relationships within it.
		
		2. **Concept Exploration**
		- Consider implicit assumptions, subtle details, underlying theories, and potential applications of the provided information.
		
		3. **Intentional Context Planning**
		- Plan how the context information can support disambiguation of the question, while not giving away the answer; ensuring the question is purposeful.
		
		4. **Detailed Assumption Expansion**
		- Consider what knowledge the question is asking about, and what information and assumptions have been made when formatting the question. Your goal is to edit the context to remove the information which would give the questions answer away to the test taker.
		
		5. **Giving Away the Answer**
		- Plan how to avoid giving away the answer in the rewritten context. 
		- Figure out what minimal set of information needs to be removed to avoid giving away the answer.
		- NEVER include the answer or information which would give away the answer in the rewritten context.
		
		### Documentation in Analysis:
		
		- Clearly document the rationale in the `<document_analysis>` tags, explaining your reasons for exclusion or inclusion decisions.
		- Clearly document what elements of the context need to be modified. What steps need to be taken and what information needs to be include most clearly and concisely (with minimal modification) remove the answer information from the context. 
		- Clearly document what information needs to be avoided in the rewritten context to prevent giving away the answer. For example if the question asks about what year a person was born, the context should not include birthday in the biographical details.
		
		
		## Context Rewriting Guidelines
		
		## Output Structure
		
		Present your final output strictly adhering the `<output_format>` tags.
		<output_format>
		[ Rewritten Context ]
		</output_format>
		
		## Output
		
		Begin by thoughtfully analyzing the provided question, answer and context within `<document_analysis>` tags. Then present the resulting edited context within `<output_format>` tags.
		
		## Important Notes
		
		- NEVER modify what the question is asking about. NEVER modify the answer. The knowledge being evaluated SHALL NOT change. 
		- Each "thought_process" should reflect careful consideration and reasoning behind your response.
		- NEVER include the answer in the rewritten context.
		- ONLY minimally modify the context as required to remove the answer information. The modified context should be as similar to the original as possible, with the answer information removed. 
		- ONLY remove answer information, do not add new information, and do not remove extraneous information.
		- Ensure rigorous adherence to output formatting and generate a single `<output_format>` tag block.
		
		
		
		<question>{question}</question>
		<answer>{answer}</answer>
		<context>{context}</context>
		"""
	\end{lstlisting}
}

\section{Answer Explanation Validation Prompt}
\label{apx:answer_validation}

{\scriptsize 
	\begin{lstlisting}[language=json]
		EXPLANATION_VALIDATION_PROMPT = """
		## Your Role
		
		You are an expert evaluator of educational content. Your goal is to produce meaningful, insightful knowledge about domain expert evaluations designed to determine competence and knowledge. 
		
		## Input Structure
		
		Your input consists of:
		
		<question>
		[A question to be answered.]
		</question>
		
		<answer>
		[The student's answer to the question.]
		</answer>
		
		<explanation>
		[An explanation for why the answer is correct.]
		</explanation>
		
		<context>
		[The text segment containing information relevant to the question.]
		</context>
		
		## Primary Objective
		
		You will be evaluating and judging the whether the student's answer and their explanation of why their answer is correct makes sense and is logically valid.
		
		Your goal is to judge whether the information presented in `<answer>` is in fact the correct answer to the `<question>` given the information in the `<context>` and whether the `<explanation>` for why the answer is correct is valid. The information in `<context>` and `<question>` can be assumed true, only the context of `<answer>` needs to be validated for correctness.
		
		### Metrics
		
		1. **Answer Correctness:** Rate from 1 to 10 how correct the provided student answer is given the information in the `<question>` and `<context>`. A rating of 1 indicates the answer is incorrect. A rating of 10 indicates the answer is correct and complete. 
		
		2. **Explanation Validity:** Rate from 1 to 10 how valid the students `<explanation>` of their answer is. The `<explanation>` should explain their thinking and the information used to determine the correct answer given the context and question. Low ratings indicate the explanation is not valid, correct, or that there is some flaw in the thinking or logic of the student. High ratings indicate the explanation is valid, correct, and explains why the answer is what it is. 
		
		## Analysis Phase
		
		Conduct careful analysis within `<document_analysis>` tags, following these steps:
		
		1. **Thoughtful Content Examination**
		- Carefully analyze the given context, identifying central ideas, nuanced themes, and significant relationships within it.
		
		2. **Concept Exploration**
		- Consider implicit assumptions, subtle details, underlying theories, and potential applications of the provided information.
		
		## Output Structure
		
		Present your final output strictly adhering the `<output_format>` tags.
		<output_format>
		Answer Correctness: [ Correctness Rating. Respond with a number in [1, 2, 3, 4, 5, 6, 7, 8, 9, 10] ]
		Explanation Validity: [ Validity Rating. Respond with a number in [1, 2, 3, 4, 5, 6, 7, 8, 9, 10] ]
		</output_format>
		
		## Output
		
		Begin by thoughtfully analyzing the provided context within `<document_analysis>` tags. Then present the resulting formatted question answer pair clearly within `<output_format>` tags.
		
		## Important Notes
		
		- Each "thought_process" should reflect careful consideration and reasoning behind your ratings.
		- Ensure rigorous adherence to output formatting.
		
		
		<question>{question}</question>
		<answer>{answer}</answer>
		<explanation>{explanation}</explanation>
		<context>{context}</context>
		"""
	\end{lstlisting}
}

\section{Question Property Validation Prompt}
\label{apx:question_validation}

{\scriptsize 
	\begin{lstlisting}[language=json]
		PROPERTIES_PROMPT = """
		## Your Role
		
		You are an expert evaluator of educational content. Your goal is to produce meaningful, insightful knowledge about domain expert evaluations designed to determine competence and knowledge. 
		
		## Input Structure
		
		Your input consists of:
		
		<question>
		[A question to be answered.]
		</question>
		
		<answer>
		[The correct answer to the question.]
		</answer>
		
		<context>
		[The text segment containing information relevant to the question.]
		</context>
		
		## Primary Objective
		
		You will be evaluating and judging the quality of test and evaluation questions across a variety of metrics. Your goal is to judge and evaluate the quality of various test and evaluation questions across a variety of metrics. The `<question>` and `<answer>` pair is grounded and drawn from the `<context>`. 
		
		### Metrics
		
		1. **Clarity:** Rate from 1 to 10 the clarity and comprehensibility (how understandable it is) of the provided `<question>`. A rating of 1 is unclear and cannot be understood or cannot be understood without the `<context>`. A rating of 10 is used for questions that are self contained, understandable, and coherent (even if the topic is complex and difficult). Questions that are missing information required to understand what is being asked rate a 1. "As of the 2015 NFL season, how many Super Bowl titles had the Denver Broncos won?" is a 10. "What event in 1861 contributed to the temporary strength of republicanism in Britain during Queen Victoria's reign?" is a 10. "In which year was the country not a member of FIFA, as indicated in the table?" is a 1. "As of the census of 2000, how many families were residing in the city?" is a 1.
		
		2. **Difficulty:** Rate form 1 to 10 the difficulty of the `<question>`. A rating of 10 is reserved for questions which require a deep understanding of the question and what is being asked by a professional domain expert. 
		
		3. **Groundedness:** Rate form 1 to 10 how grounded the provided `<question>` is in the `<context>`. A rating of 10 requires the question and answer information can found within the `<context>`. A rating of 1 indicates the question and answer information is not present in the `<context>`. This metric is only concerned with information found in the `<context>`, not outside information. The more outside information (not contained in the `<context>`) that is required to answer the question, the lower the rating.
		
		4. **Answer Give Away:** Rate from 1 to 10 how much the provided `<answer>` is given away by information in the `<question>`. A rating of 1 indicates that the information requried to answer the question is not present in the question itself. A rating of 10 indicates that the information required to answer the question is present in the question.
		
		## Analysis Phase
		
		Conduct careful analysis within `<document_analysis>` tags, following these steps:
		
		1. **Thoughtful Content Examination**
		- Carefully analyze the given context, identifying central ideas, nuanced themes, and significant relationships within it.
		
		2. **Concept Exploration**
		- Consider implicit assumptions, subtle details, underlying theories, and potential applications of the provided information.
		
		## Output Structure
		
		Present your final output strictly adhering the `<output_format>` tags.
		<output_format>
		Clarity: [ Clarity Rating (one of [1, 2, 3, 4, 5, 6, 7, 8, 9, 10] ) ]
		Difficulty: [ Difficulty Rating (one of [1, 2, 3, 4, 5, 6, 7, 8, 9, 10] ) ]
		Groundedness: [ Groundedness Rating (one of [1, 2, 3, 4, 5, 6, 7, 8, 9, 10] ) ]
		Answer Giveaway: [ Answer Giveaway Rating (one of [1, 2, 3, 4, 5, 6, 7, 8, 9, 10] ) ]
		</output_format>
		
		## Output
		
		Begin by thoughtfully analyzing the provided context within `<document_analysis>` tags. Then present the resulting formatted question answer pair clearly within `<output_format>` tags.
		
		## Important Notes
		
		- Each "thought_process" should reflect careful consideration and reasoning behind your ratings.
		- Ensure rigorous adherence to output formatting.
		
		
		<question>{question}</question>
		<answer>{answer}</answer>
		<context>{context}</context>
		"""
	\end{lstlisting}
}\textbf{}
\end{appendices}

\end{document}